\pdfoutput=1
\documentclass{article}
\usepackage{xcolor}
\usepackage[final]{corl_2026}
\makeatletter
\renewcommand{\@noticestring}{Pretrain to Adapt Workshop at the 10th Conference on Robot Learning (CoRL 2026), Austin, Texas, USA.}
\makeatother
\usepackage{times}

\input{./preamble}              %
\newcommand{\R}{\mathbb{R}}

\providecommand{\given}{\,\vert\,}
\DeclarePairedDelimiterX{\Set}[1]\{\}{\setargs{#1}}
\newcommand{\setargs}[1]{\renewcommand{\given}{\,\delimsize\vert\,}#1}

\newcommand{\Dcal}{\mathcal{D}}

\newcommand{\obshist}{\mathbf{o}}                  %
\newcommand{\actchunk}{\mathbf{a}}                 %
\newcommand{\Tobs}{T_o}                            %
\newcommand{\Tact}{T_a}                            %
\newcommand{\dobs}{d_o}                            %
\newcommand{\dact}{d_a}                            %
\newcommand{\condent}{\bar H}                      %

\newcommand{\state}{s}                             %

\newcommand{\policy}{\pi}

\newcommand{\eg}{e.g.,\ }
\newcommand{\ie}{i.e.,\ }

\usepackage{xspace}
\definecolor{armSlidingWindow}{HTML}{0072B2}
\newcommand{\armSlidingWindow}{\textcolor{armSlidingWindow}{\texttt{sliding window}}\xspace}
\definecolor{armEarlyStopping}{HTML}{3C8DC5}
\newcommand{\armEarlyStopping}{\textcolor{armEarlyStopping}{\texttt{early stopping}}\xspace}
\definecolor{armMoreSearch}{HTML}{003F7D}
\newcommand{\armMoreSearch}{\textcolor{armMoreSearch}{\texttt{more search}}\xspace}
\definecolor{armSIME}{HTML}{D55E00}
\newcommand{\armSIME}{\textcolor{armSIME}{\texttt{SIME}}\xspace}
\definecolor{armPostBC}{HTML}{B5628F}
\newcommand{\armPostBC}{\textcolor{armPostBC}{\texttt{PostBC}}\xspace}
\definecolor{armVine}{HTML}{009E73}
\newcommand{\armVine}{\textcolor{armVine}{\texttt{vine}}\xspace}
\hypersetup{
  pdftitle={Explicitly capturing demonstrator multi-modality for distillation into diffusion policies},
  pdfauthor={Shao Yuan Chew Chia, Rebecca H. Jiang, Abhinav Agarwal, Andreas Hofmann, Leslie Pack Kaelbling, Russ Tedrake},
  pdfsubject={Pretrain to Adapt Workshop at the 10th Conference on Robot Learning (CoRL 2026)},
}
\usepackage{xspace}

\newcommand{\piE}{\policy_E}                 %
\newcommand{\pitheta}{\policy_\theta}        %
\newcommand{\Phat}{\hat{P}_k}                %
\newcommand{\Hgen}{\mathcal{H}^{\mathrm{gen}}_0}
\newcommand{\Hmem}{\mathcal{H}^{\mathrm{mem}}_0}

\begin{document}

\title{Explicitly capturing demonstrator multi-modality for distillation into diffusion policies}

\author{%
Shao Yuan Chew Chia$^{1}$, Rebecca H. Jiang$^{1}$, Abhinav Agarwal$^{1}$\\
\bf Andreas Hofmann$^{2}$, Leslie Pack Kaelbling$^{1}$, Russ Tedrake$^{1}$ \\
$^{1}$Computer Science and Artificial Intelligence Laboratory (CSAIL), \\
Massachusetts Institute of Technology, Cambridge, MA, USA \\
$^{2}$Pickle Robot Company, Charlestown, MA, USA \\
\texttt{shaoyuan@mit.edu}
}
\maketitle
\vspace{-14pt}

\begin{abstract}
\looseness = -1 
The \textit{diversity} of a distilled diffusion policy's outputs matters beyond how it impacts the policy's success rate.
In particular, downstream applications like test-time search and policy improvement require exploration.
Thus, we want to distill multi-modal demonstrators into diffusion policies which reproduce their conditional action distributions.
Standard data collection for behavior cloning applies a sliding window to demonstration episodes,
which pairs \textit{unique observation histories} with actions.
These datasets can be fit by a policy that is either multi-modal, or unimodal with high Lipschitz constant to capture a diversity of actions seen at \textit{nearby} observation samples.
Prior work finds that diffusion policies trained on sparse data learn the latter, memorizing one action per observation.
We adapt \textit{vine sampling} to behavior cloning data collection to resolve this ambiguity: at some states the demonstrator visits, query it $k$ times from \textit{exactly} the same observation.
On a synthetic task, we study how the number of actions per observation, $k$, affects the learned policy, and find that larger $k$ yields a policy that better matches the demonstrator's conditional distribution, including its multi-modality and diversity.
On a 2D truck unloading task, vine sampling likewise leads to a policy closer to the demonstrator and substantially improves both best-of-$N$ and expert iteration performance.
\end{abstract}

\keywords{behavior cloning, data collection, multi-modal action distributions}

\section{Introduction}
\label{sec:introduction}
\begin{figure}[!b]
  \centering
  \IfFileExists{figures/truck_qualitative.pdf}%
    {\includegraphics[width=\linewidth]{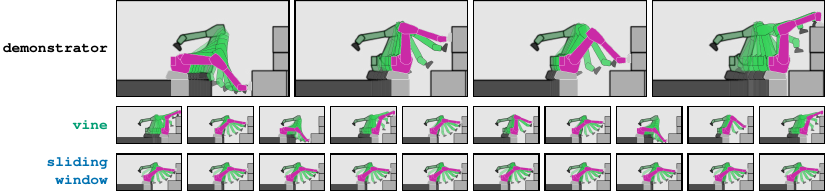}}%
    {\framebox[\linewidth]{\rule{0pt}{1.3in}\itshape placeholder: truck\_qualitative}}
  \caption{Actions from the truck unloading demonstrator of \cref{sec:truck} (top), the \armVine policy (middle), and the \armSlidingWindow policy (bottom), queried at the same state of a held-out problem instance. Each panel shows the observation image with future arm poses overlaid; poses with suction on are shown in pink. Of 10 samples each, the \armVine policy's reach three of the four grasps, while the \armSlidingWindow policy's all approach the same grasp.}
  \label{fig:truck_qualitative}
\end{figure}

\looseness = -1
Diffusion and flow policies trained with behavior cloning (BC) have proven effective on a range of challenging manipulation tasks \citep{black2024pi0, trilbm2026}.
In BC, a policy is trained to imitate an \textit{expert demonstrator}, such as a human teleoperator, a motion planner, or a scripted controller.
In many applications, however, success rate is not the only metric of interest.
The diversity of a policy's action distribution, and in particular its \textit{multi-modality} -- that is, its capacity to take different actions in response to the same observation -- is important when the policy is used for exploration by some downstream procedure.
For instance, test-time search draws many candidate actions from a policy and selects the best with a verifier or value function \citep{qiInferenceTimeEnhancementGenerative2025, kwok2025robomonkey}, and is limited when the policy's samples are insufficiently diverse \citep{dalir2026reranking}.
Expert iteration \citep{anthony2017thinking} and reinforcement learning (RL) post-training \citep{wagenmaker2025postbc} similarly shift a policy toward higher-reward actions that it samples, directly or through a search it guides.
Diversity also matters when a policy is used as a skill inside a task-and-motion-planning framework \citep{garrett2021integrated}, where constraints may only become known at test time and a unimodal policy's single approach may prove inadmissible.

For RL post-training, this need has been made precise as a coverage condition: to reach the demonstrator's performance, the pretrained policy must, in some cases, place sufficient probability mass on the demonstrator's actions \citep{wagenmaker2025postbc}.
In robotics, BC policies often violate this coverage condition \citep{wagenmaker2025postbc, hong2026tmrl}.
Even when the demonstrator is stochastic and multi-modal, as in the top row of \cref{fig:truck_qualitative}, the trained policies often concentrate on a single mode per observation, as in its bottom row.

\begin{wrapfigure}{R}{0.5\textwidth}
  \vspace{-\intextsep}
  \centering
  \includegraphics[width=\linewidth]{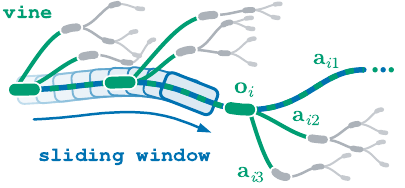}
  \caption{The choices that a demonstrator has the capacity to take branch along an episode. Sliding window datasets record one action per observation history, while vine datasets record several counterfactual actions at a random subset of observation histories. Gray actions are recorded in neither dataset.}
  \label{fig:vine_tree}
  \vspace{-\intextsep}
\end{wrapfigure}
\looseness = -1 These low-diversity policies follow from the way training datasets are constructed.
Standard practice collects full episodes of the demonstrator solving the task, then applies a \textit{sliding window} that, at each timestep, pairs an observation history with the demonstrator's actions over the prediction horizon \citep{chi2023diffusionpolicy}.
Each episode records a particular sequence of actions taken by the demonstrator. At each decision point, the demonstrator could have chosen differently, but the resulting dataset does not reflect these counterfactual actions (\cref{fig:vine_tree}).
Because observations are often high dimensional and continuous, and reached states stochastic, other episodes never repeat exact observation histories.\footnote{When the observation space is discrete and small, the same observation history may repeat across episodes, and the sliding-window dataset may contain explicit multi-modality.}
The demonstrator's multi-modality is therefore not explicitly represented in the data \citep{pan2026muchado}.
At the same time, diffusion models at these dataset scales, given sufficiently expressive architectures and enough training, operate in or near the memorization regime (\cref{sec:related-work}).
The standard pipeline therefore memorizes a dataset in which each observation is paired with a single mode; low policy diversity is an expected outcome.
Some architectures may be better than others at nonetheless learning to generalize based on nearby observations paired with different action modes.  Yet, these datasets themselves remain fundamentally ambiguous: the underlying demonstrator could be interpreted as multi-modal or unimodal with high Lipschitz constant. In this paper, we focus on disambiguating the dataset.

The natural solution is to record multiple samples per condition, and some methods already do so \citep{carvalho2023mpd,dong2026flow2one}. In these works, however, it is one component of a larger method; its effect has not been isolated.
To our knowledge, no work in BC has varied the number of recorded actions per observation, fixing dataset size and training budget, and studied the relationship between the conditional distributions of the demonstrator, dataset, and trained policy.
We study this relationship in a controlled synthetic setting (\cref{sec:analysis}) and in a simulated 2D truck unloading problem that captures challenges relevant to real robot tasks.
Furthermore, we study how this intervention affects performance on our ultimate goal of downstream policy self-improvement.

We consider \textit{re-queryable settings}, in which the demonstrator can be queried multiple times from the same world state.
Such settings arise in simulation, where the environment can be reset between queries, with closed-loop demonstrators such as a human teleoperator or a feedback controller.
They also arise, in simulation or the real world, with demonstrators that compute their actions without changing the state of the world, such as an open-loop motion planner.

This paper makes two \textbf{contributions}.

\looseness = -1 \textbf{Analysis.} In a controlled synthetic setting, we study how the number of \textit{samples} (recorded actions) per \textit{condition} (observation) $k$ affects how well the policy learns the demonstrator's conditional distribution at matched total dataset size (\cref{sec:analysis}).
With more samples per condition, the model still memorizes, but the memorized samples better capture the diversity and multi-modality of the demonstrator.

\textbf{Application to BC for robotics.} Motivated by this analysis, we adapt the vine procedure of \citet{schulman2015trpo} into a BC data-collection protocol that explicitly captures the demonstrator's multi-modality: at each step, with some probability, re-query the stochastic demonstrator $k$ times from the current state (\cref{sec:approach}).
On a simulated 2D truck unloading task, at matched dataset size, the resulting policy better matches the demonstrator's action distribution and diversity, and substantially improves both best-of-$N$ search and expert iteration (\cref{sec:results}).

\section{Related Work}
\label{sec:related-work}

\textbf{Memorization in conditional diffusion models.}
The exact minimizer of the denoising objective is the score of the Gaussian-smoothed empirical distribution of the training set, so a sufficiently expressive and fully trained diffusion model memorizes, generating only its training samples \citep{baptista2025memorization, ye2025provable}.
In the conditional setting, this holds at each training condition: the model generates the uniform distribution over the samples stored there \citep{baptista2025memorization}.
Finer conditioning granularity, with unique per-sample labels as the extreme case, accelerates memorization \citep{gu2025memorization}.
A sliding-window dataset, in which every condition is a unique observation history, is an analogous case for BC, and diffusion policies in the sparse-data regime indeed behave like retrieval over their training set \citep{he2026demystifying}.

\textbf{Increasing policy diversity downstream of data collection.}
Low sample diversity in behavior-cloned generative policies hampers self-improvement \citep{jin2025sime}, reranking-based search \citep{dalir2026reranking}, and RL fine-tuning \citep{wagenmaker2025postbc, hong2026tmrl}.
\citet{wagenmaker2025postbc} and \citet{hong2026tmrl} synthesize increased diversity at training time, by noising actions and conditioning respectively, and \citet{jin2025sime,soe2025onmanifold} at inference time, with perturbations to conditioning or in a learned latent space.
\citet{pan2026muchado} find that generative policies trained on standard BC benchmarks show little multi-modality and do not owe their single-rollout success to it; consistent with our analysis, they attribute its absence to data sparsity, as the same policies fit both modes when the data are dense, but they do not measure downstream search or improvement.
\citet{belkhale2023data} likewise note that individual states are visited only a few times, so a policy cannot match a high-entropy expert there, and curate demonstrations toward lower expert entropy to improve single-rollout success.
In contrast, we retain the expert's entropy for downstream search by recording more samples per condition, which in our setting increases single-rollout success.

\textbf{Multiple samples per condition.}
Training on multiple samples per condition is common where a condition has more than one correct answer: language models are fine-tuned on multiple responses per prompt, with gains that grow with the number of responses \citep{yuan2023scaling, modelottery2026}, and segmentation models are trained on labels from multiple annotators \citep{kohl2018probabilistic}.
In robotics, grasp planners are trained on multiple grasps per object \citep{mousavian2019graspnet, urain2023se3df}, and motion planners are trained on several trajectories per planning problem to learn multi-modal priors \citep{urain2022implicit,carvalho2023mpd}.
Most closely related to our work,
\citet{dong2026flow2one} distill a multi-step flow policy into a one-step student. They sample the teacher several times per observation and train the student with a set-level Implicit Maximum Likelihood Estimation objective that preserves the teacher's modes,
reporting that success rate rises with the number of samples per observation.
We study this core idea for diffusion models trained with the standard denoising objective to assess its achievement of two goals: reproducing the demonstrator's conditional action distribution, and facilitating downstream search and self-improvement.

\section{Problem Formulation}
\label{sec:problem-formulation}

\subsection{Setting}
\label{sec:setting}

\looseness = -1 The policy conditions on an observation history $\obshist \in \R^{\Tobs \times \dobs}$ and outputs actions $\actchunk \in \R^{\Tact \times \dact}$, where 
$\Tobs$ is the length of the history, 
$\Tact$ the prediction horizon (number of timesteps the predicted actions span), and 
$\dobs$ and $\dact$ are the dimensions of a single timestep's observation and action, respectively.
Square brackets index within an object: $\actchunk[h, c]$ is the $c$-th action coordinate, $h$ timesteps into the future.

\looseness=-1
We model the expert demonstrator as a conditional distribution $\piE(\cdot \given \obshist)$ over actions.
BC trains a policy $\pitheta$ on a dataset of \textit{training samples}: observation--action pairs $(\obshist, \actchunk)$ with $\actchunk \sim \piE(\cdot \given \obshist)$.
In our setting, multiple actions may share the same observation history; we write the dataset grouped by it,
\(
  \Dcal = \big\{\, (\obshist_i,\ \actchunk_{i1}, \dots, \actchunk_{ik}) \,\big\}_{i=1}^{n},
\)
with $n$ distinct conditions and $k$ actions per condition, each drawn independently from $\piE(\cdot \given \obshist_i)$.
This is equivalent to a dataset of $n \cdot k$ pairs $\{(\obshist_i, \actchunk_{ij})\}$.
Standard practice takes $k = 1$; we analyze the effect of increasing $k$, holding the dataset size $n \cdot k$ fixed (\cref{fig:dataset_sweep}).
We assume the policy class is expressive enough to represent multi-modal action distributions for the task.

\subsection{Study of the trained policy}
\label{sec:objective}

\looseness = -1 We are interested in whether $\pitheta(\cdot \given \obshist)$ reproduces $\piE(\cdot \given \obshist)$ as a distribution, rather than in the success rate of a single rollout of $\pitheta$.
We evaluate this via two metrics.
The policy's conditional should be close to the demonstrator's as a distribution, which we capture via the $1$-Wasserstein distance $W_1$ in \eqref{eq:decomp}.
It should also place enough mass on each of the demonstrator's modes that $N$ draws, as in best-of-$N$ search, reach them. $W_1$ does not capture this, so we measure it as mode recall@$N$ (\cref{sec:analysis_results}).

At a training condition $\obshist_i$, we compare three distributions over actions:
the demonstrator's conditional $\piE(\cdot \given \obshist_i)$ is the object we want the policy to reproduce;
the empirical conditional $\Phat(\obshist_i) = \frac1k \sum_{j=1}^k \delta_{\actchunk_{ij}}$ is the uniform distribution over the $k$ training actions that the collection protocol stored at that condition; and
the trained policy's conditional is $\pitheta(\cdot \given \obshist_i)$.
\citet{baptista2025memorization} show that the minimizer of the conditional denoising objective at $\obshist_i$ is the empirical score, the score of $\Phat(\obshist_i)$ under the forward noising process, and that sampling with it reproduces $\Phat(\obshist_i)$ exactly.
Because $W_1$ is a metric, the triangle inequality with $\Phat(\obshist_i)$ as a waypoint gives
\begin{equation}
  \underbrace{W_1\big(\pitheta,\,\piE\big)}_{\mathclap{\text{(LHS) policy error}}}
  \hspace{1.5em}\le\hspace{1.5em}
  \underbrace{W_1\big(\pitheta,\,\Phat\big)}_{\mathclap{\text{(A) memorization error}}}
  \hspace{1.5em}+\hspace{1.5em}
  \underbrace{W_1\big(\Phat,\,\piE\big)}_{\mathclap{\text{(B) data-collection sampling error}}} ,
  \label{eq:decomp}
\end{equation}
where all three distributions are conditioned on $\obshist_i$.
Term (A) is a property of the model: how far its conditional is from the training actions at that observation.
Term (B) is a property of data collection: how well $k$ draws represent the distribution they came from.
Taking $\pitheta$ as the waypoint instead gives $W_1(\Phat,\piE) \le W_1(\pitheta,\Phat) + W_1(\pitheta,\piE)$, which bounds policy error from below by (B) $-$ (A).
The policy error therefore lies within  $\pm$ (A) of (B); when (A) is small, as it is near the minimizer described above, small (B) is necessary and sufficient for small policy error.
Thus, a policy that memorizes its training actions can reduce its policy error only through a smaller (B), which on average requires a larger $k$.
This motivates studying the isolated effects of increasing $k$ on policy error and performance.

\section{Analysis}
\label{sec:analysis}

\subsection{Synthetic task}
\label{sec:synth_task}

\begin{figure}[t]
  \centering
  \includegraphics[width=\linewidth]{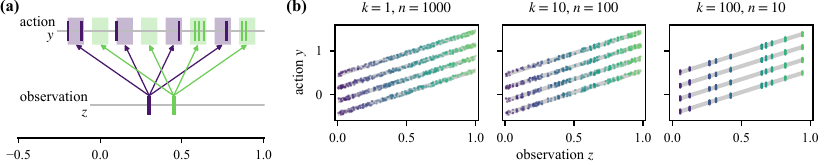}
  \caption{(a) Coordinate $y$ of the synthetic task at two observations, with $k = 5$ actions sampled at each; arrows run from each observation to its four valid modes. 
  (b) Visualization of synthetic datasets with fixed total dataset size $n\cdot k = 1000$, but varied numbers of observations $n$, and actions per observations $k$; gray diagonals show the valid modes of the task. At fixed dataset size, more actions per condition comes at the cost of sparsity in $z$.}
  \label{fig:two_rail_and_dataset_sweep}
  \sublabel{fig:two_rail}
  \sublabel{fig:dataset_sweep}
\end{figure}

\looseness = -1 The synthetic task is defined on observations $\obshist \in \R^{1 \times 16}$ and actions $\actchunk \in \R^{1 \times 16}$: 16-dimensional observations and actions, with no notion of time.
The observation has one informative dimension, denoted $z := \obshist[1, 1]$, and 15 irrelevant ``nuisance'' dimensions.
The observation marginal is $\mu = \mathrm{Unif}[0,1]^{16}$, so $z \sim \mathrm{Unif}[0,1]$.
The nuisance dimensions enlarge the observation space, distancing the 16D observations from one another, to mimic the sparse-observation property seen in diffusion policies with high-dimensional image observation histories.
As in these real observations, only a small subset of the high-dimensional input is decision-relevant.
The \textit{primary} action coordinate $y := \actchunk[1, 1]$ determines whether an action is valid: it must lie in one of four intervals shifted by the observation $z$ (\cref{fig:two_rail}).
The remaining 15 coordinates can take any value, analogous to redundant degrees of freedom, as when inverse kinematics admits multiple joint configurations for one end-effector pose.

The demonstrator's conditional distribution $\piE(\cdot \given \obshist)$ is given by
\begin{equation}
  \label{eq:synthetic_task}
  y = z + u, \quad u \sim p_R, \qquad 
  \actchunk[1, c] \;=\; \varepsilon_c \quad (c = 2, \dots, 16),
  \quad \varepsilon_c \overset{\text{i.i.d.}}{\sim} \mathcal{N}(0, 0.3^2),
 \end{equation}
where the residual distribution $p_R$ is a mixture of the four valid modes, equally weighted and uniform (\cref{fig:two_rail}).
The other 15 action dimensions are unconditional Gaussians. Because they are random, a policy draw can lie close to a training action in all 16 dimensions only if the policy reproduces that action, which our retrieval metric relies on.

\subsection{Findings}
\label{sec:analysis_results}
\begin{figure}[t]
  \centering
  \includegraphics[width=\linewidth]{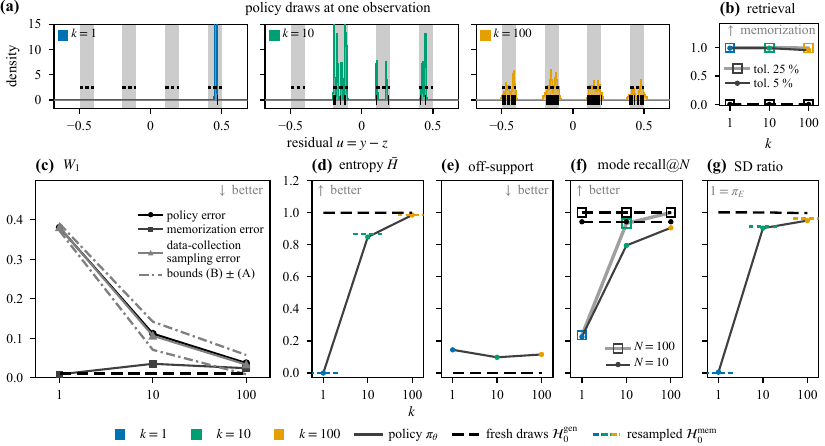}
  \caption{Policies trained on the three datasets of \cref{fig:dataset_sweep}, queried at their training observations.  
  (a) Each policy's draws at a representative training observation.
  Dashed segments and gray bands are $\pi_E$ and its support.
  Short black lines mark the training actions at each observation.
  (b)--(g) The metrics of \cref{sec:analysis_results}, averaged over training observations; (c) shows the terms of \eqref{eq:decomp} on the primary coordinate $y$.
  Reference lines are fresh draws from the demonstrator (black dashed, $\Hgen$) and resampled draws from the training actions (colored dashed, $\Hmem$).
  }
  \label{fig:analysis_train}
  \sublabel{fig:analysis_train_residual_law}
  \sublabel{fig:analysis_train_retrieval}
  \sublabel{fig:analysis_train_decomp}
  \sublabel{fig:analysis_train_entropy}
  \sublabel{fig:analysis_train_off_support}
  \sublabel{fig:analysis_train_recall}
  \sublabel{fig:analysis_train_primary_sd}
\end{figure}

We run a sweep of $k \in \{1, 10, 100\}$ at matched budget $n \cdot k = 1000$, so the number of conditions is $n \in \{1000, 100, 10\}$ (\cref{fig:dataset_sweep}).
Every metric is computed from 1000 policy draws per condition, averaged over conditions.
To give a qualitative sense of what the model learned, we plot a histogram of the residual $u = y - z$ of the policy's primary-task action when the policy is queried repeatedly at a particular training condition (\cref{fig:analysis_train_residual_law}).
Each policy's draws closely match the distribution of training actions seen at each condition.
For $k=1$, since there is only a single action in the dataset for this condition,
the other three valid modes are not represented.

\looseness=-1 
To test whether a policy memorizes its training data, we compare each of its draws with the nearest training action, 
as in nearest-neighbour tests of memorization for image diffusion models \citep{carlini2023extracting, gu2025memorization} and diffusion policies \citep{he2026demystifying}.
\textit{Retrieval} is the fraction of draws whose 16D distance to that action is at most 5\% or 25\% of the mean distance between two demonstrator draws at the same condition (\cref{fig:analysis_train_retrieval}).
All three policies show high retrieval at both the 5\% and 25\% levels,
whereas essentially none of the fresh draws from the demonstrator fall within these tolerances of the training actions.

Plotting the $W_1$ distances between $\piE$, $\pitheta$, and $\Phat$ on the primary coordinate $y$ (\cref{fig:analysis_train_decomp}),
the data-collection sampling error (B) decreases as $k$ increases, \ie the dataset better represents the demonstrator's conditional distribution.
The memorization error (A) is small at every $k$, consistent with the retrieval results.
Since the policy error lies within (A) of (B), it closely tracks (B) and falls with it.

To understand the policy's multi-modal behavior,
we estimate two quantities.  The first is conditional behavior entropy $\condent(\obshist)$ \citep{jia2024d3il}, the entropy of the policy's distribution over the demonstrator's modes at a condition, normalized to $[0,1]$, which measures how evenly probability mass is distributed among those modes.
The second is
off-support mass: the proportion of draws that land outside the demonstrator's support.
The entropy $\condent(\obshist)$ is normalized over the mass in the demonstrator's support, so off-support mass complements it.
The entropy $\condent$ increases with $k$, from 0 at $k=1$ (all on-support mass lies in one mode) to 1 at $k=100$ (mass evenly distributed across the modes) (\cref{fig:analysis_train_entropy}), while the off-support mass is similar across $k$ values (\cref{fig:analysis_train_off_support}).

Mode recall@$N$ is the expected proportion of demonstrator mass lying in modes that at least one of $N$ independent policy draws lands in.
Mode recall@$N$ is also substantially higher at larger $k$ (\cref{fig:analysis_train_recall}).
As expected, at $k=1$, increasing $N$ from 10 to 100 barely moves recall; more sampling at inference time does not recover the demonstrator's multi-modality.
The primary-task standard-deviation (SD) ratio is the SD of the policy's primary-task draws divided by the demonstrator's; it is $1$ in expectation for the demonstrator and $0$ for a collapsed policy.
On this measure of diversity, the $k=10$ and $k=100$ policies also match the demonstrator much more closely than the $k=1$ policy does (\cref{fig:analysis_train_primary_sd}).

At held-out conditions, compared to the $k=1$ policy, the $k=10$ and $k=100$ policies continue to have favorable policy error, entropy $\condent$, mode recall and SD ratio at every distance $d$ from the query to its nearest training observation (\cref{fig:analysis_held_out}).
The $k=1$ policy remains a point mass at every distance, while the $k=10$ and $k=100$ policies gradually lose their four-mode structure, so the gaps narrow with distance. 
Retrieval falls for all policies, fastest for $k=1$.

\looseness = -1 With enough data, the $k = 1$ policy also reaches a low policy error.
At $n \cdot k = 10^5$, it matches the $k = 10$ and $k = 100$ policies in policy error, entropy $\condent$, off-support mass and SD ratio (\cref{fig:analysis_train_nk_sweep}).
Its retrieval there is near zero (\cref{fig:analysis_train_nk_retrieval}), and its policy error is far below the sampling error (B) of a $k = 1$ dataset (\cref{fig:analysis_train_decomp}), whose mean does not depend on $n$.
By the lower bound of \cref{sec:objective}, it can reach that error only by moving away from its training actions, which it does by generalizing across conditions.
For this task and model architecture, the advantage of larger $k$ is therefore limited to the sparse-data regime, where a model with enough capacity and training memorizes.
Diffusion policies at typical BC dataset scales operate in or near this regime (\cref{sec:related-work}), which motivates collecting several actions per condition (\cref{sec:approach}).

\section{Data collection method: vine sampling}
\label{sec:approach}
\looseness=-1
Motivated by the preceding analysis, we investigate the effects of extending $k>1$ data collection to robot BC. To facilitate this, we adapt \textit{vine sampling},\footnote{We adapt the vine procedure of \citet{schulman2015trpo}
(also used by \citet{kazemnejad2025vineppo} and \citet{hamid2026polychromic}),
which branches policy rollouts to estimate advantages for policy-gradient updates;
we instead use the demonstrator's branched actions as BC training data.}
(\cref{alg:vine} in \cref{sec:appendix_vine}).
The demonstrator $\piE$ visits states as usual: replanning periodically and enacting the stored plan in between.
At each timestep, with probability $p$, we query the expert $k$ times from the current state, resetting the environment between queries if the state is modified by the action decision process (e.g., simulation for closed-loop control).
Only these states enter the dataset, each carrying exactly $k$ actions.
Because the expert is multi-modal, the $k$ actions at a labeled state demonstrate up to $k$ different ways of continuing from it; they capture the demonstrator's conditional distribution in a way the sliding window 
approach does not (\cref{sec:introduction,fig:vine_tree}).
\Cref{sec:appendix_vine} discusses other sampling strategies, and how to apply vine sampling when the demonstrator cannot be queried from some visited states.

\section{Truck unloading experiments}
\label{sec:results}
In this section, we compare policies trained with our proposed vine sampling data-collection approach and standard sliding window on a 2D truck unloading task (pictured in \cref{fig:truck_qualitative}).
In this task, a planar robot with a suction gripper removes four boxes from a truck by placing each box on a conveyor belt that moves with its base.
Success is defined as removing all boxes before the timeout at 80s.  We also measure throughput, in pieces per hour.
The task captures several challenges of real-world robotics, including meaningful multi-modality: the robot has to make both discrete (which box to grasp) and continuous (how to grasp it) decisions.
At each decision, the number of modes grows with the number of boxes; over an episode the number of possible pick orders grows combinatorially.
We describe the task and the model-based demonstrator we implement in more detail in \cref{sec:truck}.

Each method collects demonstrations on the same 2,000 problem instances.
For the $k = 1$ (\armSlidingWindow) dataset we apply the standard sliding window to the episodes. The $k = 10$ (\armVine) dataset is collected with \cref{alg:vine} at $p = 0.1$.
Since $p \cdot k = 1$, the two methods produce the same expected number of training samples per episode. Both datasets are truncated to 370,000 training samples, so the collection budget $n \cdot k$ is matched.
Like in the synthetic experiment (\cref{sec:analysis}), at matched dataset size, the vine dataset captures explicit multi-modality at the cost of being sparser in conditioning space.
We then train one policy on each dataset for 100,000 steps with the diffusion policy architecture of \citet{chi2023diffusionpolicy}.
We compare against four baselines that use the sliding-window dataset and intervene downstream of data collection, at training or inference time: \armEarlyStopping, \armMoreSearch, \armSIME and \armPostBC, each described with its results below.

We evaluate the trained policies on two downstream use cases: test-time best-of-$N$ search, and expert iteration \citep{anthony2017thinking} policy improvement, where we use best-of-$N$ in place of Monte-Carlo Tree Search.
Their implementations are described in \cref{sec:appendix_truck_search}.
We report results for $N \in \{10, 100\}$ and the raw policy, which executes predicted actions without search.
We initialize expert iteration with the \armSlidingWindow, \armVine and \armSIME policies.

\begin{figure}[ht]
  \centering
  \IfFileExists{figures/truck_results.pdf}%
    {\includegraphics[width=\linewidth]{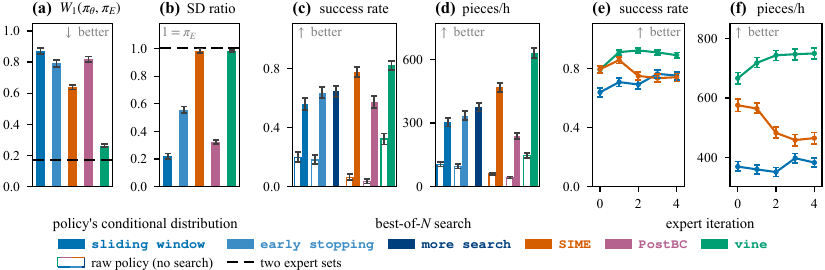}}%
    {\framebox[\linewidth]{\rule{0pt}{1.5in}\itshape placeholder: truck\_results}}
  \caption{Truck unloading experiment. 
  (a), (b) Policy error $W_1(\pitheta,\piE)$ and SD ratio per method at held-out conditions. 
  (c), (d) Success rate and throughput, in best-of-$N$ search; $N=10$ for solid bars except \armMoreSearch where $N=100$. 
  (e), (f) The same two metrics over expert iteration for the \armSlidingWindow, \armSIME, and \armVine methods.
  Plotted values are means and error bars are 95\% intervals.
  }
  \label{fig:truck_results}
  \sublabel{fig:truck_policy_error}
  \sublabel{fig:truck_sd_ratio}
  \sublabel{fig:truck_bon_success}
  \sublabel{fig:truck_bon_pph}
  \sublabel{fig:truck_exit_success}
  \sublabel{fig:truck_exit_pph}
\end{figure}

\looseness = -1 \textbf{\armVine and \armSlidingWindow.} As anticipated, \armVine shows superior policy error and SD ratio, suggesting that it captures the demonstrator's underlying distribution better than alternatives. 
It also demonstrates the intended impacts on search, yielding high success rates and throughput.  Furthermore, these metrics improve via expert iteration, with success rate falling slightly at later iterations, which we attribute to imperfect distillation between iterations combined with diminishing diversity at later iterations, as only the base policy is trained via vine sampling.  
\armSlidingWindow's success rate rises, but its successful episodes take longer on average, so throughput changes little. We attribute this to its limited diversity: with similar candidate actions, best-of-$N$ search has little better behavior to select and distill. In contrast, \armVine completes episodes faster over the course of expert iteration.

\textbf{\armEarlyStopping.}
SD ratio falls as training progresses and the policy memorizes (\cref{fig:truck_steps_sd_ratio}). We use an earlier checkpoint as a simple attempt to maintain diversity, at the cost of reduced training.
We choose the 20k checkpoint, whose raw policy success rate and throughput are similar to those of the 100k \armSlidingWindow checkpoint.  Indeed, this yields a higher SD ratio, and thereby improves best-of-$N$ search.  However, while behaviorally diverse, it does not mimic the demonstrator's choices as well as \armVine (higher policy error), leading to worse downstream search than 
\armVine.

\textbf{\armMoreSearch.}
We evaluate best-of-$N$ with standard sliding window and $N=100$ instead of the $N=10$ used with all other methods.  This increases success rate and throughput, but performance is still far from that of \armVine.
If the policy has a collapsed distribution, sampling repeatedly from it is ineffective at recovering the demonstrator's coverage --
we see this effect clearly in \cref{sec:analysis_results}, \cref{fig:analysis_train_recall}.

\textbf{\armSIME.}
At inference time, \armSIME \citep{jin2025sime} adds noise to the output of the observation encoder, annealed across the denoising timesteps of the diffusion policy.
We choose the SD of the added noise such that \armSIME matches the demonstrator's SD (\cref{fig:truck_sime_sd_ratio}).  
In search, \armSIME approaches the success rate of \armVine, but falls short on throughput, suggesting that the form of injected randomness adds enough options for the search to find an admissible, albeit inefficient, solution.
As a raw policy, \armSIME produces more failures relative to \armSlidingWindow, which shares its checkpoint.
We attribute this to the noise injected into the observation encoding, which reduces the precision of the policy's actions, such as those needed to avoid contact with other objects.
Across expert iteration, \armSIME's success rate and throughput fall in later iterations and its motions become increasingly jerky (\cref{fig:truck_exit_success,fig:truck_exit_pph}), presumably because these imprecise actions are distilled into the policy at each iteration.

\looseness = -1 \textbf{\armPostBC.}
\armPostBC \citep{wagenmaker2025postbc} adds Gaussian noise to each training action, with the covariance of an ensemble's predictions of the mean action at that observation.
This widens the policy's distribution, but does not add the demonstrator's counterfactual actions, which \armVine captures.
Accordingly, \armPostBC raises SD ratio and lowers policy error only slightly relative to \armSlidingWindow, and its raw policy succeeds far less often.

We conclude that, on this task, adding diversity to a policy trained on a sliding-window dataset, at either training or inference time, is a poor substitute for recording the demonstrator's multi-modality in the dataset itself.
\armVine, whose dataset records that multi-modality explicitly, performs best in both search and policy self-improvement.

\section{Conclusions and limitations}
\label{sec:conclusions}
In this work, we study recording several demonstrator actions per observation as a way to distill a multi-modal demonstrator into a diffusion policy.
In a controlled synthetic setting at a matched dataset size, the policy memorizes its training data at every $k$, but with larger $k$ the memorized data contain more of the demonstrator's diversity and multi-modality.
Motivated by this finding, we apply vine sampling to BC data collection, which queries the demonstrator $k$ times at a random subset of the states the demonstrator visits.
On a 2D truck unloading task, the vine policy reproduces the demonstrator's conditional distribution more closely than the sliding-window policy, and substantially improves both best-of-$N$ search and expert iteration.

Vine sampling can require more demonstrator queries than the sliding window for the same number of training samples: the sliding window turns every timestep of an executed episode into a training sample, while vine sampling spends the same queries to complete the episode, and one additional query per training sample.
For our truck unloading demonstrator, which plans once per pick and place, collecting the vine dataset costs about 30 times the compute of the sliding-window dataset at matched size.
For conciseness, we have not investigated optimal values for the re-query probability $p$ and the query count $k$. 
Furthermore, the best $k$ for best-of-$N$ search may not be best for rollouts of the raw policy.
Finally, we study vine sampling only in simulation, where every method can be evaluated on the same problem instances and at a scale that would be costly on hardware.
Our implementations of best-of-$N$ search and expert iteration also use the simulator to evaluate candidate actions.
When the demonstrator does not change the state of the world when queried, vine sampling does not require environment reset and can be run on hardware, but the re-queried plans are then never executed; their quality is not verified.
Our synthetic analysis suggests that the advantage of larger $k$ is limited to the sparse-data regime (\cref{sec:analysis_results}). Future work could study whether large robot foundation models \citep{black2024pi0, trilbm2026}, at their data scales, model sizes and training recipes, operate in this regime, and hence whether vine sampling would improve the policies that test-time search and policy improvement build on.

\subsubsection*{Acknowledgments}

This work was supported by Pickle Robot Company and The Charles Stark Draper Laboratory, Inc., where Rebecca H. Jiang is a Draper Scholar.
Part of this work was done while Shao Yuan Chew Chia was an intern at Pickle Robot Company.

\bibliography{references}

\appendix
\crefalias{section}{appendix}
\crefalias{subsection}{subappendix}
\crefalias{subsubsection}{subsubappendix}

\section{Synthetic Analysis}
\label{sec:appendix_synth_details}

This appendix gives the details of the synthetic analysis of \cref{sec:analysis} and its results at held-out conditions and other dataset sizes.

\subsection{Task, datasets and training}
\label{sec:appendix_synth_setup}

Each of the four modes of the residual distribution $p_R$ of \cref{eq:synthetic_task} is an interval of width $0.1$, and neighbouring modes are $0.2$ apart, so the modes are centered at $\pm 0.15$ and $\pm 0.45$ and the support is $[-0.5, 0.5]$.

We collect twelve datasets for this synthetic task, one for each $k \in \{1, 10, 100\}$ at each total dataset size $n \cdot k \in \{10^2, 10^3, 10^4, 10^5\}$, and train a policy on each dataset.
The three datasets with $n \cdot k = 10^3$ are plotted in \cref{fig:dataset_sweep}, and their policies are evaluated on training conditions in \cref{fig:analysis_train}, and on held-out conditions in \cref{fig:analysis_held_out}.
All twelve use the architecture and training recipe in \cref{tab:appendix_synth_hparams}.

\begin{table}[!ht]
 \centering
 \small
 \begin{tabular}{ll}
 \toprule
 Implementation & LeRobot \texttt{DiffusionPolicy} \\
 Architecture & 1D convolutional U-Net, one level of width 256 \\
 Parameters & 3.1M \\
 Diffusion steps & 100 for training and sampling (DDPM) \\
 Noise schedule & cosine \\
 Prediction target & $\epsilon$ \\
 Optimizer & AdamW, learning rate $10^{-4}$, weight decay $10^{-6}$ \\
 Learning rate schedule & 500-step warmup, then constant \\
 Batch size & 256 \\
 Training steps & $5 \times 10^{6}$ \\
 Hardware & NVIDIA RTX 4090, 13.5\,h per policy \\
 \bottomrule
 \end{tabular}
 \caption{Architecture and training recipe of the twelve synthetic policies.}
 \label{tab:appendix_synth_hparams}
\end{table}

\subsection{Metrics}
\label{sec:appendix_synth_evaluation}

For evaluations of the synthetic task policies, the $W_1$ terms against $\piE$ in \cref{eq:decomp} are computed exactly from the cumulative distribution function of $p_R$, without demonstrator samples, while the distribution of $\pitheta$ is approximated with 1000 samples, and $\Phat$ is exact from the training datasets.
For retrieval, the nearest training action is searched over all $n \cdot k$ training actions, not only those stored at the query's condition.
Conditional behavior entropy is undefined when no draws land in the demonstrator's support, and is noisy when few do.
Thus, when fewer than 15 of the 1000 land in the support, we set it to $0$, its value for a collapsed policy, and keep the condition in the average. 
Dropping those conditions would mean each policy's entropy is averaged over a different set of conditions.
At $k = 1$ and $n \cdot k = 10^3$, this happens at $7\%$ of training conditions and $58\%$ of held-out queries.
At $k = 10$ and $k = 100$ this does not happen in any training conditions and occurs in under $1\%$ of held-out queries.
The reference lines $\Hgen$ and $\Hmem$ use the same number of draws as the policies (1000 draws).
Plotted values are means over up to 128 training conditions per policy, evenly spaced in the informative observation coordinate $z$.

\subsection{Held-out conditions}
\label{sec:appendix_analysis}

\begin{figure}[!ht]
  \centering
  \includegraphics[width=\linewidth]{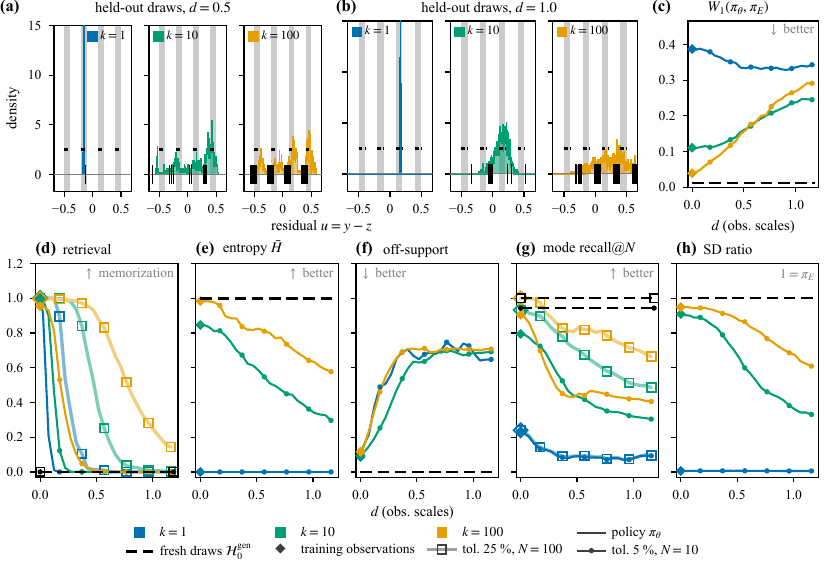}
  \caption{The policies of \cref{fig:analysis_train}, queried at held-out conditions at distance $d$ from the nearest training observation, in units of the mean distance between two training observations.
  (a), (b) Each policy's draws at a representative held-out condition at $d = 0.5$ and $d = 1.0$.
  Dashed segments and gray bands are $\piE$ and its support.
  Short black lines mark the nearest training observation's actions, shifted into the query's residual frame.
  (c)--(h) The metrics of \cref{fig:analysis_train}, averaged within bins of $d$; diamonds at $d = 0$ are the values at training observations.
  Reference lines are fresh draws from the demonstrator (black dashed, $\Hgen$).
  }
  \label{fig:analysis_held_out}
  \sublabel{fig:analysis_held_out_residual_law_half_dist}
  \sublabel{fig:analysis_held_out_residual_law_full_dist}
  \sublabel{fig:analysis_held_out_policy_error}
  \sublabel{fig:analysis_held_out_retrieval}
  \sublabel{fig:analysis_held_out_entropy}
  \sublabel{fig:analysis_held_out_off_support}
  \sublabel{fig:analysis_held_out_recall}
  \sublabel{fig:analysis_held_out_primary_sd}
\end{figure}

Held-out queries evaluated in \cref{fig:analysis_held_out} are taken along straight lines that start at training observations and run in random directions in the 16-dimensional observation space.
For each query, $d$ is the distance to its nearest training observation, in units of the mean distance between two training observations computed on the $k=1$, $n=1000$ dataset.

We split $d$ into 24 bins from $0$ to $1.18$ and draw 96 queries per bin for each policy, at most one from each line originating from a training observation.
Plotted values at held-out conditions are means over the queries in each bin.

\subsection{Results against total dataset size}
\label{sec:appendix_synth_nk}

\begin{figure}[!ht]
  \centering
  \includegraphics[width=\linewidth]{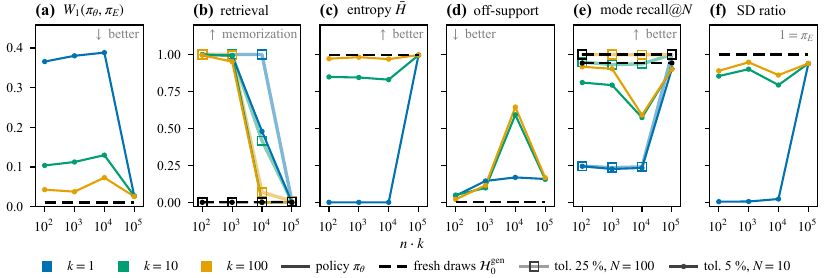}
  \caption{The metrics of \cref{fig:analysis_train} against the total dataset size $n \cdot k$, one curve per $k$, averaged over training observations.
  Reference lines are fresh draws from the demonstrator (black dashed, $\Hgen$).
  }
  \label{fig:analysis_train_nk_sweep}
  \sublabel{fig:analysis_train_nk_policy_error}
  \sublabel{fig:analysis_train_nk_retrieval}
  \sublabel{fig:analysis_train_nk_entropy}
  \sublabel{fig:analysis_train_nk_off_support}
  \sublabel{fig:analysis_train_nk_recall}
  \sublabel{fig:analysis_train_nk_primary_sd}
\end{figure}

\Cref{fig:analysis_train_nk_sweep} shows the metrics of \cref{fig:analysis_train} for all twelve policies.

\clearpage
\section{Vine Sampling}
\label{sec:appendix_vine}

This appendix gives the vine sampling procedure of \cref{sec:approach} in full (\cref{alg:vine}).

\begin{algorithm}[!ht]
\algtop
\small
\setlength{\algomargin}{2em}
\DontPrintSemicolon
\SetAlgoNoEnd
\SetKwInOut{Input}{input}
\SetKwInOut{Output}{output}
\Input{expert demonstrator $\piE$, environment, re-query probability $p$, query multiplicity $k$}
$\Dcal \gets \emptyset$\;
\For{each episode}{
  \While{episode not done}{
    form the observation history $\obshist$\;
    record the state $\state$\;
    \If{$\xi \sim \mathrm{Unif}[0,1] < p$\label{alg:line6}}{
      \For{$j = 1, \dots, k$}{
        $\actchunk_j \sim \piE(\cdot \given \obshist)$\;
        \lIf{producing $\actchunk_j$ modified $\state$}{reset the environment to $\state$}
      }
      $\Dcal \gets \Dcal \cup \{(\obshist, \actchunk_1, \dots, \actchunk_k)\}$\;
    }
    \If{replan required}{
      $\tau \sim \piE(\cdot \given \obshist)$\;
    }
    pop the leading action of $\tau$, and advance\;
  }
}
\Return{dataset $\Dcal$}
\caption{Vine sampling}
\alglabel{alg:vine}
\end{algorithm}

Besides vine sampling, other sampling strategies are possible; for example, the episode could continue branched along each of the $k$ re-queried actions rather than only one.
We forgo this, as it would shift the state distribution along the time axis, over-representing states later in the episode.
Vine sampling labels a random fraction $p$ of the states along a single trajectory, so it has $p$ times as many conditions as the sliding window, distributed along the episode in the same way.
It is simple to implement and can be carried out in real-world data collection when the demonstrator does not require resets, since it is only queried along states that are actually visited.

The algorithm, as written, requires that the demonstrator be queryable from any visited state $\state$.
In cases where this does not hold, \eg systems limited by partial observability, \cref{alg:vine} can be modified to only re-query from states that the demonstrator can plan from, simply by adding a condition at line \ref{alg:line6}.

\section{Truck Unloading}
\label{sec:appendix_truck_details}

This appendix gives the details of the truck unloading experiments of \cref{sec:results} and their baselines.

\subsection{Task and demonstrator}
\label{sec:truck}

The task is simulated in Drake~\citep{drake}.
Task failure occurs upon object inter-penetration beyond a threshold, or if the low level controller cannot track the commanded joint positions within a tolerance.
These failures are signs of poor policy performance, the former a simulator-based detection of in-collision motions.
A failed episode's throughput is calculated using the 80s timeout as the duration.
Each timestep incurs a step reward of -0.1. A reward of -100 is incurred on failure or timeout.

Each observation is an image and 6D state: the four joint positions, suction command (on/off), and a flag for whether any box is currently grasped.
The policy takes in an observation history of $\Tobs = 2$ steps and outputs an action spanning $\Tact = 32$ timesteps.
Each action consists of four joint positions and a suction command.
Each problem instance randomizes box sizes and configurations.
By keeping the task in 2D and using a side view of the truck, we remove partial observability as a confounding factor.

The demonstrator generates plans by internally generating an ordered list of candidate grasps, attempting to plan motions to them, continuing down the list until an admissible motion is found.  It randomly shuffles the order of the candidate grasps, over boxes and their faces.
The demonstrator plans motions using bidirectional rapidly-exploring random trees (BiRRT) \citep{lavalle1998rrt, kuffner2000rrtconnect} to generate initial motions which are subsequently refined via trajectory optimization.
Two queries from one state may select different boxes, different faces of a box, different arm configurations to reach a selected grasp, or different continuous motions to reach the grasp pose (\cref{fig:truck_qualitative}, top row).
The expert can be re-queried at any visited state, necessary for \cref{alg:vine}.

\subsection{Problem instances and datasets}
\label{sec:appendix_truck_task}

The policies are trained on problem instances 0 to 1{,}999, and expert iteration collects demonstrations on instances 2{,}000 to 6{,}999.
All other evaluations use the same 600 test instances, separate from both.
On the 600 test instances, the demonstrator succeeds in $95.8\%$ of episodes with a throughput of 766 pieces per hour.

\label{sec:appendix_truck_datasets}
Episodes in which the demonstrator failed are kept in both datasets.
It succeeded in 1{,}899 of the 2{,}000 episodes of the sliding-window dataset and 1{,}878 of the vine dataset's.
The vine dataset stores the plans returned by the re-queries, while the sliding-window dataset stores the executed trajectory, which differs from the plan by the controller's tracking error.
The expert re-queries used to measure policy error (\cref{sec:appendix_truck_e0}) are also plans, so they are the same kind of action as the vine policy's training actions.
Per episode, collecting the vine dataset took 55 seconds of a 96-core machine and the sliding-window dataset took 1.8 seconds, the roughly 30-fold cost noted in \cref{sec:conclusions}.

\subsection{Policy and value networks}
\label{sec:appendix_truck_training}

The \armSlidingWindow and \armVine policies use the settings in \cref{tab:truck_policy_hparams} and differ only in their dataset.
\armPostBC differs only in its training targets (\cref{sec:appendix_truck_baselines}).

\begin{table}[!ht]
\centering
\caption{Diffusion policy settings.}
\label{tab:truck_policy_hparams}
\small
\begin{tabular}{ll}
\toprule
Architecture & CNN-based U-Net of \citet{chi2023diffusionpolicy}, channels $(256, 512, 1024)$ \\
Image encoder & ResNet-18 trained from scratch, GroupNorm, spatial softmax \\
Diffusion steps & 100 for training and sampling (DDPM) \\
Noise schedule & cosine \\
Optimizer & AdamW, learning rate $3 \times 10^{-4}$ (constant), weight decay $10^{-6}$ \\
Batch size & 2048 \\
Training steps & 100{,}000 \\
EMA of weights & yes \\
Hardware & two NVIDIA H100 GPUs, about 8\,h per policy \\
\bottomrule
\end{tabular}
\end{table}

\Cref{tab:truck_value_hparams} gives the value network's settings.

\begin{table}[!ht]
\centering
\caption{Value network settings.}
\label{tab:truck_value_hparams}
\small
\begin{tabular}{ll}
\toprule
Input & one image and the 6D state \\
Image encoder & ResNet-18 trained from scratch \\
State encoder & MLP \\
Head & MLP with a $\tanh$ output \\
Optimizer & Adam, learning rate $2 \times 10^{-3}$ \\
Batch size & 1024 \\
Training steps & 100{,}000 \\
Hardware & one NVIDIA H100 GPU, 4.9\,h \\
\bottomrule
\end{tabular}
\end{table}

\subsection{Metrics}
\label{sec:appendix_truck_e0}

Policy error and SD ratio (\cref{fig:truck_policy_error,fig:truck_sd_ratio}) compare a policy's actions with the demonstrator's at the same states.
We choose these states along demonstrator episodes on the 600 test problem instances, taking the current state with probability $0.005$ at each timestep, as in \cref{alg:vine}.
At each of the resulting 576 states we query the demonstrator 200 times.
Some queries fail to return a plan, so we keep only the 507 states where all 200 succeed, thus every evaluated state has the same number of demonstrator actions.

We split the 200 demonstrator actions at each state into two sets of 100, which we call the estimate set and the reference set.
The estimate set is our estimate of $\piE$ at that state.
The reference set shows what a policy that matched the demonstrator exactly would score with 100 draws, and gives the reference lines in the figures.
We draw 100 actions from each policy at each state to compute each of the metrics.

Each action spans 32 timesteps of four joint positions and a suction command, which have different units.
To compute the distance between two actions, we first divide each action dimension $c$ by the demonstrator's standard deviation $\varsigma_c$ in that dimension, computed over every timestep of all demonstrator actions at the 507 states, and take the root-mean-square difference over timesteps and dimensions:
\begin{equation}
 d(\actchunk, \actchunk') \;=\; \sqrt{\frac{1}{\Tact \dact} \sum_{h=1}^{\Tact} \sum_{c=1}^{\dact} \left( \frac{\actchunk[h, c] - \actchunk'[h, c]}{\varsigma_c} \right)^{\!2}}.
 \label{eq:truck_chunk_metric}
\end{equation}
Policy error at a state is $W_1$ under this distance between the 100 policy draws and the estimate set.
For the SD ratio, we measure the spread of a set of actions as their root-mean-square distance from their mean, and divide the spread of the 100 policy draws by the spread of the estimate set, to get a ratio.
Both metrics are averaged over the 507 states.
Their reference lines are computed the same way, with the reference set in place of the policy draws.

All intervals in \cref{fig:truck_results} are 95\% intervals.
For policy error, SD ratio and throughput, they are bootstrap confidence intervals.
We resample problem instances with replacement 1000 times, adding all states in each drawn instance to the mean, since states from the same problem instance may not be independent.
For success rate, they are equal-tailed Bayesian credible intervals of the $\mathrm{Beta}(1 + \text{successes}, 1 + \text{failures})$ posterior.

\subsection{Best-of-$N$ search and expert iteration}
\label{sec:appendix_truck_search}

Our implementation of best-of-$N$ search draws $N$ actions from the policy, simulates the leading 20 timesteps of each, discards candidates whose simulations fail, ranks the rest by the simulated reward plus a value network's estimate at the terminal state, and executes the best.
The value network is trained on the same 2,000 demonstrator episodes used for policy training, to predict the reward-to-go given the current observation, and is shared across all methods.
If every candidate fails in simulation, the episode fails.
Planning time does not advance the simulation clock, so a slow search does not lower throughput.

In each iteration of expert iteration, a policy and a value network drive a best-of-100 search to collect demonstrations from 1{,}000 fresh problem instances.
The search can be viewed as a policy improvement operator that generates demonstrations that are stronger than the raw policy's.
Both the policy and value networks are then fine-tuned on this data, enabling the expert in the subsequent iteration to learn from these new best-of-$N$-improved demonstrations.
In \armSIME's run, SIME's noise is added in every iteration, with its standard deviation fixed at its value for the base policy.
Within the expert iteration process, we apply the standard sliding-window collection for all three runs, so they otherwise differ only in the base policy that they were initialized with.
While each method's expert iteration is initialized with a shared value function, the value function is fine-tuned during expert iteration, diverging across the methods.
All three runs collect demonstrations on the same instances.
In each iteration, the policy and the value network are fine-tuned for 10{,}000 steps, and data from $j$ iterations earlier are replayed with weight $0.8^{j}$.
We do not evaluate the policies separately during expert iteration.
\Cref{fig:truck_exit_success,fig:truck_exit_pph} plot each policy's results on the 1{,}000 instances it collected, so iteration 0 is the base policy under best-of-100 on new instances.

\subsection{SIME and PostBC}
\label{sec:appendix_truck_baselines}

\begin{figure}[!ht]
  \centering
  \IfFileExists{figures/truck_sime_sweep.pdf}%
    {\includegraphics[width=\linewidth]{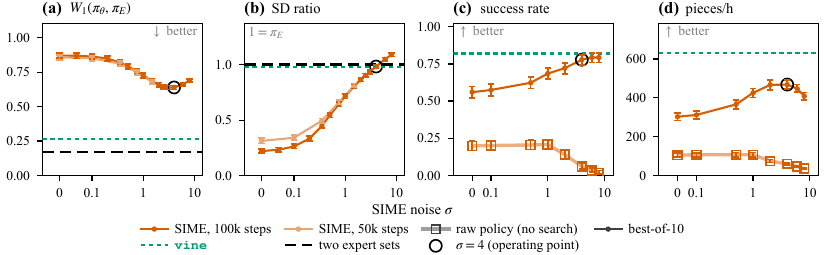}}%
    {\framebox[\linewidth]{\rule{0pt}{1.5in}\itshape placeholder: truck\_sime\_sweep}}
  \caption{The \armSIME baseline's noise sweep on the \armSlidingWindow policy, against the noise scale $\sigma$; $\sigma = 0$ is the unmodified policy.
  (a) Policy error and (b) SD ratio at the states of \cref{fig:truck_results}, for the 100{,}000-step checkpoint (dark) and the 50{,}000-step one (light).
  (c) Success rate and (d) throughput on the 600 test instances at the 100{,}000-step checkpoint, under best-of-10 (filled) and for the raw policy (open).
  Reference lines are two independent sets of expert re-queries (black dashed) and the \armVine policy at 100{,}000 steps, under best-of-10 in (c) and (d) (green).
  The circled point is $\sigma = 4$, the value used in \cref{fig:truck_results}.
  Means, with 95\% intervals as in \cref{fig:truck_results}.
  }
  \label{fig:truck_sime_sweep}
  \sublabel{fig:truck_sime_policy_error}
  \sublabel{fig:truck_sime_sd_ratio}
  \sublabel{fig:truck_sime_bon_success}
  \sublabel{fig:truck_sime_bon_pph}
\end{figure}

At inference-time, \armSIME \citep{jin2025sime} adds noise to the output of the observation encoder, annealed across the denoising timesteps of the diffusion policy.
We apply it to the 100{,}000-step \armSlidingWindow policy without retraining.
The noise has standard deviation $\sigma s_{\mathrm{img}}$, where $s_{\mathrm{img}}$ is the root-mean-square standard deviation of the observation encoder's image output over the 576 states.
It is scaled by a factor that falls linearly from one at the first denoising step to zero at the last, and each action draw keeps one noise sample across all steps.
This follows the SIME paper.
We use $\sigma = 4$, where the SD ratio is closest to one and policy error is lowest (\cref{fig:truck_sime_sweep}).

Where BC fits the empirical distribution of the dataset, \armPostBC \citep{wagenmaker2025postbc} fits a posterior predictive estimate of the demonstrator's conditional distribution $\piE$.
This distribution is wider than the empirical one where the dataset is sparse in conditioning space, and approaches it as data scales.
In practice, \armPostBC trains an ensemble of regressors on bootstrap resamples of the demonstrations, then adds Gaussian noise to each training action, with the covariance of the ensemble's predictions at that observation.
Each regressor predicts the conditional mean action, so the noise represents uncertainty about the mean.
The ensemble has 10 regressors.
Each is a copy of the policy's observation encoder followed by an MLP with three hidden layers of 512 units, trained on a bootstrap resample of the 2{,}000 episodes for 6{,}000 steps at batch size 256.
The noise covariance spans the whole $32 \times 5$ action and is not rescaled ($\alpha = 1$ in the notation of \citet{wagenmaker2025postbc}), and new noise is drawn each time a training sample is used.
We use their default ensemble size and $\alpha$ without tuning, but train the ensemble for fewer epochs than they do since our 2,000 episodes are far larger than their 5–300-trajectory datasets.

\subsection{Results against training step}
\label{sec:appendix_truck}

\begin{figure}[!ht]
  \centering
  \IfFileExists{figures/truck_steps.pdf}%
    {\includegraphics[width=\linewidth]{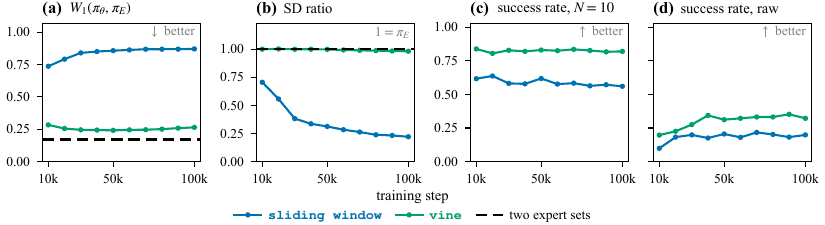}}%
    {\framebox[\linewidth]{\rule{0pt}{1.5in}\itshape placeholder: truck\_steps}}
  \caption{The \armSlidingWindow and \armVine policies against training step.
  (a) Policy error and (b) SD ratio at the states of \cref{fig:truck_results}.
  (c) Success rate under best-of-10 and (d) of the raw policy, on the 600 test instances.
  Reference lines in (a) and (b) are two independent sets of expert re-queries (black dashed).
  }
  \label{fig:truck_steps}
  \sublabel{fig:truck_steps_policy_error}
  \sublabel{fig:truck_steps_sd_ratio}
  \sublabel{fig:truck_steps_bon_success}
  \sublabel{fig:truck_steps_raw_success}
\end{figure}

\Cref{fig:truck_steps} shows the \armSlidingWindow and \armVine policies at every 10{,}000-step checkpoint.

\end{document}